\documentclass[preprint,authoryear]{elsarticle}
\usepackage[breaklinks=true]{hyperref}
\usepackage{inputenc}
\usepackage{csquotes}
\usepackage{epsfig}
\usepackage{amsmath}
\usepackage{amssymb}
\usepackage{siunitx}
\usepackage{amsfonts}
\usepackage{booktabs}
\usepackage{multirow}
\usepackage{caption}
\usepackage{subcaption}
\usepackage{algorithm2e}
\usepackage{textcomp}
\usepackage{xspace}
\usepackage{multimedia}
\usepackage{graphicx}
\usepackage{float}
\usepackage{wrapfig}
\usepackage{pgfplots}
\pgfplotsset{compat=1.17}
\usetikzlibrary{spy}
\usepackage[toc,page]{appendix}
\usepackage[english]{babel}
\usepackage{overpic}
\usepackage{color,soul}

\DeclareMathOperator*{\argmin}{arg\,min}

\newcommand{\revb}{}
\newcommand{\revi}{}

\begin{document}
% \doclicenseThis

\allowdisplaybreaks

\title{\revi{Overcoming the Distance Estimation Bottleneck in Estimating Animal Abundance with Camera Traps}}

\author[1]{Timm Haucke}
\ead{haucke@cs.uni-bonn.de}
\author[2]{Hjalmar S. Kühl}
\author[2]{Jacqueline Hoyer}
\author[1]{Volker Steinhage}
\ead{steinhage@cs.uni-bonn.de}

\date{December 15, 2021}

\address[1]{University of Bonn, Institute of Computer Science IV, Friedrich-Hirzebruch-Allee 8, Bonn 53115, Germany}
\address[2]{German Centre for Integrative Biodiversity Research (iDiv) Halle-Jena-Leipzig, Puschstrasse 4, 04103 Leipzig, Germany}

\begin{abstract}
    \noindent
    \revb{
    % TODO: first sentence
    The biodiversity crisis is still accelerating, despite increasing efforts by the international community. Estimating animal abundance is of critical importance to assess, for example, the consequences of land-use change and invasive species on community composition, or the effectiveness of conservation interventions.
    Various approaches have been developed to estimate abundance of unmarked animal populations. Whereas these approaches differ in methodological details, they all require the estimation of the effective area surveyed in front of a camera trap. Until now camera-to-animal distance measurements are derived by laborious, manual and subjective estimation methods. To overcome this distance estimation bottleneck, this study proposes an automatized pipeline utilizing monocular depth estimation and depth image calibration methods. We are able to reduce the manual effort required by a factor greater than $21$ and provide our system at \href{https://timm.haucke.xyz/publications/distance-estimation-animal-abundance}{https://timm.haucke.xyz/publications/distance-estimation-animal-abundance}
    }
\end{abstract}


\maketitle
\let\thefootnote\relax\footnotetext{
\\
© 2021. This manuscript version is made available under the CC-BY-NC-ND 4.0 license \href{https://creativecommons.org/licenses/by-nc-nd/4.0/}{https://creativecommons.org/licenses/by-nc-nd/4.0/}
}

\clearpage

%\tableofcontents
%\clearpage

\section{Introduction}\label{sec:introduction}
The dramatic decrease in biodiversity and wild animal populations require the accurate and large-scale monitoring of wildlife. \revb{Camera trapping has become a widely used approach for surveying wildlife populations \citep{steenweg2017scaling}. Animal abundance can be estimated from camera trap footage using capture-recapture methods which require the individual identification of animals \citep{o2011camera}. This is, however, challenging with species that do not have unique individual markings. Therefore, a number of methods have been developed for the estimation of abundance of unmarked animal populations that do not require identification of individuals \citep{gilbert2021abundance}. These include the random encounter model (REM) \citep{rowcliffe2008estimating}, the random encounter and staying time model (REST) \citep{nakashima2018estimating}, the time-to-event model (TTE), space-to-event model (STE), instantaneous estimator (IS) \citep{moeller2018three} and camera trap distance sampling \citep{camera_trap_distance_sampling}.}

\subsection{Problem statement: laborious distance estimation}\label{sec:existing_methods}
\revb{Whereas the approaches differ in methodological details \citep{palencia2021assessing,gilbert2021abundance}, they all have in common that an estimate of the effective area surveyed by a camera trap is needed. This is essential in order to relate the number of animal observations to a measure of spatial survey effort. The effective area surveyed is derived by the opening angle of the camera and the effective detection distance. The effective detection distance is the distance below which as many individuals are missed as are seen beyond \citep{hofmeester2017simple}. With increasing distance from a camera trap, the detection probability of animals decreases due to occlusion. Not accounting for detection probability and thus animals not seen, would lead to biased estimates of the effective detection distance and thus the effective area surveyed. Effective detection distances generally require that camera-to-animal observation distances or distances to some objects in the detection zone can be derived.}
%
% TODO
\revb{However, currently all deployed camera traps are monocular\footnote{Camera traps including depth estimation are currently just subject to research on wildlife monitoring \citep{haucke2021exploiting}.}, recording images or video clips using a single lens at a time. These monocular images and videos do not deliver distance information in a direct way.}
\revb{
% One approach would be to use reference objects permanently in the camera's field of view. Natural reference objects such as trees, rocks etc. are not available in every scenario and do not mark, due to their size, no sharp position. Additionally natural objects can be occluded over time (a rock occluded by growing plants) or may disappear (a tree broken by a storm). Artificial reference objects show the same problems with respect to occlusion and additionally may have invasive effects on animal behavior. Therefore, related work shows two prominent methods for estimating such depth information based on monocular camera trap imagery.
Related work shows two prominent methods for estimating such depth information based on monocular camera trap imagery.

\textbf{Visual estimation by reference objects:} distances between the camera trap's lens and the midpoint of each detected animal are estimated by comparing animal locations in recorded video or picture to reference objects placed at known distances from the camera trap (e.g., from $\SI{1}{\metre}$ to $\SI{12}{\metre}$ in $\SI{1}{\metre}$ intervals). Reference objects can either be imaged only once when the camera traps are installed \citep{camera_trap_distance_sampling} or be placed permanently in the scene \citep{palencia2021assessing} to be visible in each observation. Permanently placed reference objects might seem generally preferable, as their position can be compared with animals in the same respective image under ideal conditions. However, they might get obstructed by snow, growing plants or other objects, again requiring comparing different images (one with the animal and another with the reference object unobstructed). Either way, comparing the locations of animals and reference objects is not only very laborious but can also be subjective.

\textbf{On-site distance measurement:} distances between the camera trap location and each previously observed animal are measured in the field, e.g., using a measuring tape \citep{tape_measurement} or an ultrasonic distance sensor such the Vertex IV system of \citet{vertex} \citep{vertex_measurement}. The on-site distance  measurement is slightly less subjective than visual estimation by reference images, but even more laborious, since the location of the camera trap has to be visited in person to obtain measurements for each animal observation.
}
\noindent

% TODO:
% Obviously, the visual estimation by reference images relies heavily on the \revi{person's capability} to estimate the position and distance of an observed animal just by comparing its visual appearance in the field image with the reference images. While some animal locations can be quite reliably estimated (e.g., when the animal is positioned between two trees \revi{of} known distance), other locations are more difficult to estimate (e.g. when the animal is located in high grass and its position on the ground is not visible).

\subsection{Contribution: automating distance estimation}\label{sec:goals}
\begin{figure}
    \centering
    \includegraphics[width=0.75\textwidth]{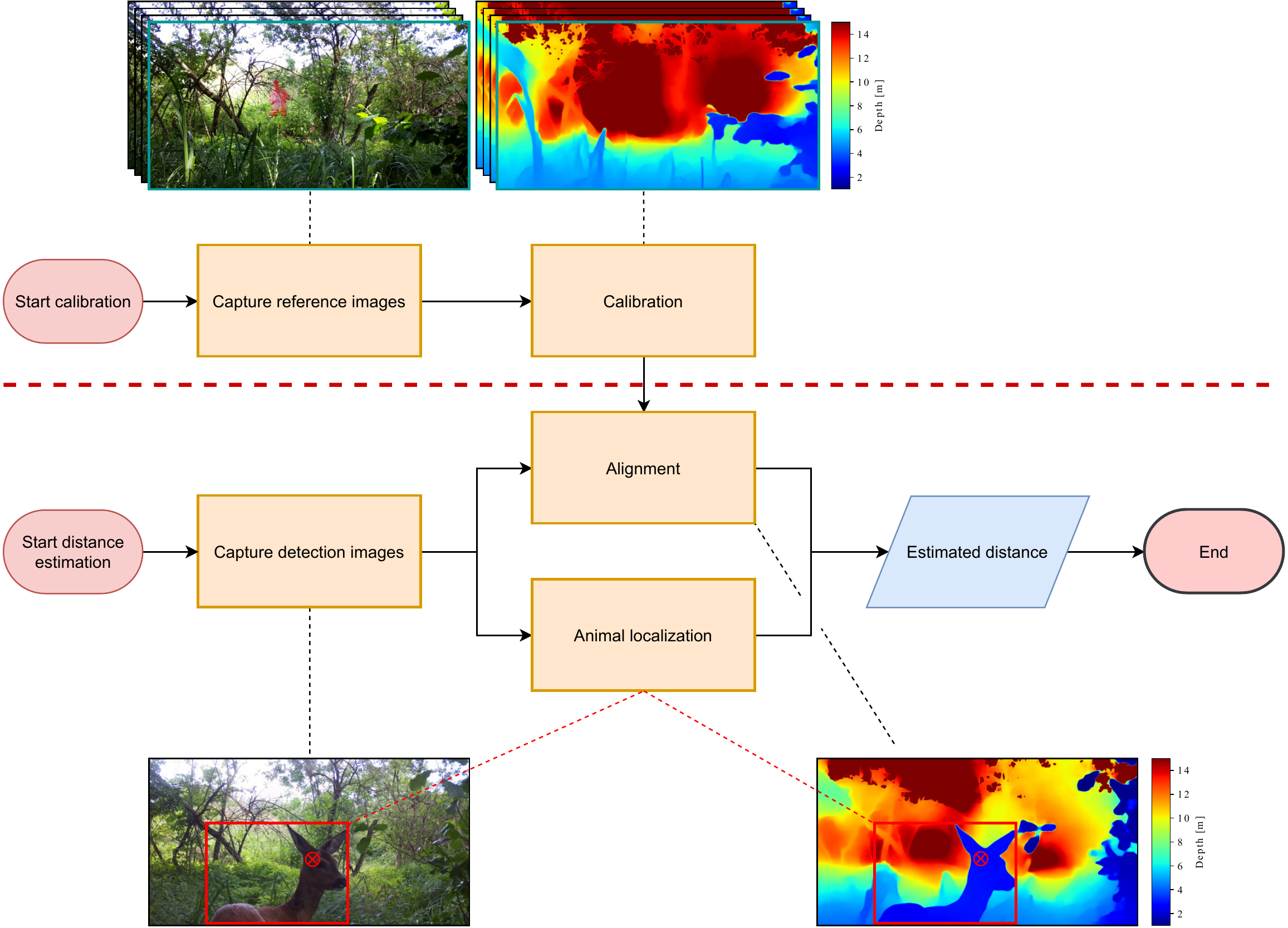}
    \caption{\revb{The overall pipeline consists of the calibration workflow and the distance estimation workflow. The calibration workflow is depicted in the upper area and derives the calibrated target depth image (highlighted in blue) of the observed transect based on a number $N$ of reference images showing landmarks placed in different distances. It is explained and visualized in more detail in section \ref{sec:calibration} and fig. \ref{fig:reference_examples}, respectively. 
    The distance estimation workflow estimates the real distance of an observed animal based on the localization of the animal in the observation image and an adjustment by the calibrated depth image of the transect. It is explained and visualized in more detail in section \ref{sec:localization} and fig. \ref{fig:detection_examples}, respectively.}}
    \label{fig:workflow}
\end{figure}

% TODO feinschliff
\revb{
In this study, we propose a two-step pipeline to automate the estimation of camera-to-animal distances from monocular camera images. (1) The \textbf{calibration workflow} delivers the automated calibration of the observed transect using reference images and measurements. (2) The \textbf{distance estimation workflow} employs the calibration of the observed transect to automatically estimate camera-to-animal distances in camera trap images showing observed animals. Figure \ref{fig:workflow} depicts in the upper area the calibration workflow and in the lower area the distance estimation workflow.
}

The \textbf{calibration workflow} starts with annotated reference images of the transect. The annotation of a reference image depicts the exact distance between the camera and a visible landmark (i.e., a distinct object placed on the transect with just the exact distance to the camera). Generally, several reference images are captured with landmarks placed at different distances, e.g., from 1 to 12 m. The calibration workflow generates from these given annotated reference images of a transect a calibrated depth image of the transect with exact distance measurements given in meters. These calibrated depth images are visualized as heatmaps where the distance is lowest in blue and highest in red. This calibration workflow is explained in more detail in section \ref{sec:calibration}.

The \textbf{distance estimation workflow} starts with a so-called observation image, i.e., an image showing an animal observed in the transect. %For such an observation image, uncalibrated distances are estimated via monocular depth estimation, i.e., distances are only estimated relatively without exact specification in meters. By alignment of such an uncalibrated distance image with the calibrated depth image of the transect (delivered by the calibration workflow) exact estimations of the camera-animal distance in meters are derived. This distance estimation workflow is explained in more detail in section \ref{sec:animal_distance_estimation}.
Using the calibrated depth image of the transect (delivered by the calibration workflow), an exact estimation of the camera-animal distance in meters is derived. This distance estimation workflow is explained in more detail in section \ref{sec:animal_distance_estimation}.

%\subsection{Challenges}\label{sec:challenges}
%Such estimated depth images are not directly usable for acquiring metric distances, as the scale information is unreliable, i.e. only the relative distances are estimated with high accuracy. It is therefore crucial to calibrate the estimated depth images using known scale information. Additionally, the scale of estimated depth images often shifts, even between images of the same scene. Therefore, depth images of some scene not only have to be calibrated using known scale information but also aligned among themselves as to maintain a common scale and obtain usable results. The last challenge we face is the localization of the animal in the congruent color and depth images, which we tackle by a combination of species-agnostic deep learning detection models, class activation maps (CAMs, \cite{cam}) and depth histograms.

\section{Data material}
\label{sec:data}
\begin{figure}[H]
     \centering
     \begin{subfigure}[b]{0.49\textwidth}
         \centering
         \includegraphics[width=\textwidth]{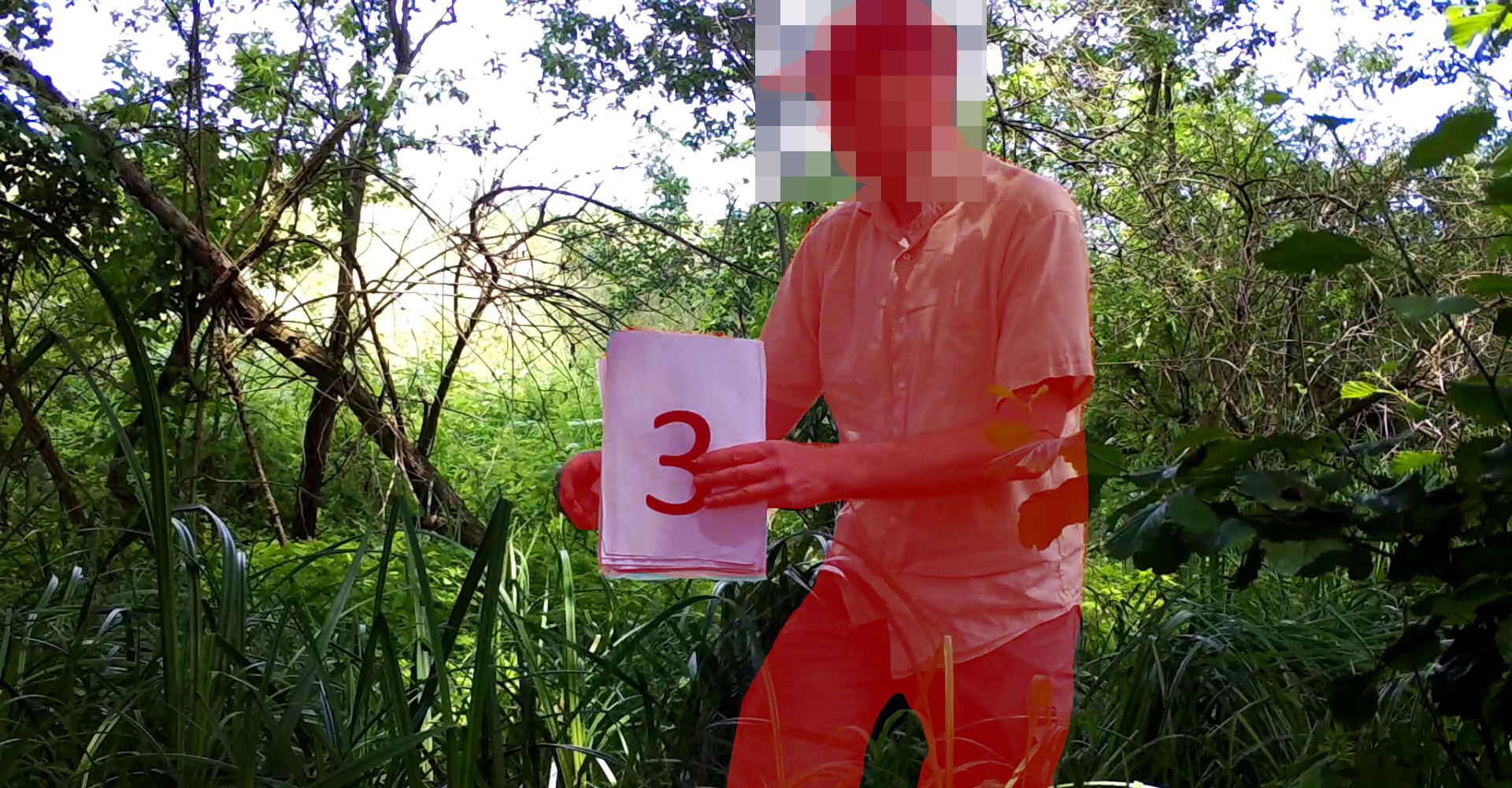}
         \caption{Reference image at $\SI{3}{\metre}$}
         \label{fig:reference_example_1}
     \end{subfigure}
     \hfill
     \begin{subfigure}[b]{0.49\textwidth}
         \centering
         \includegraphics[width=\textwidth]{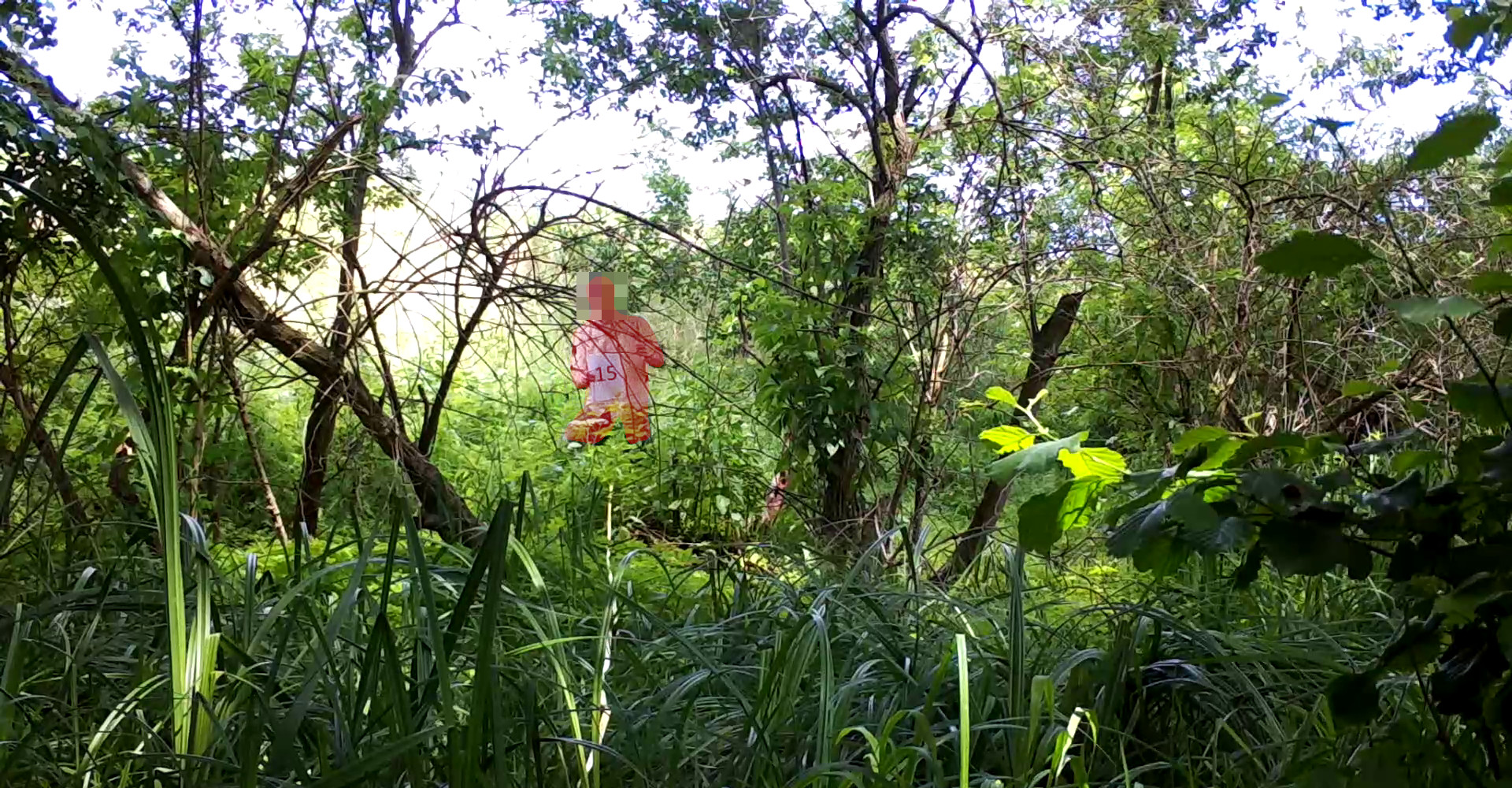}
         \caption{Reference image at $\SI{15}{\metre}$}
         \label{fig:reference_example_2}
     \end{subfigure}
     \hfill
    \caption{Examples of two reference images with the researcher and the paper sheet acting as a landmark positioned in a distance of 3 meters and 15 meters with respect to the camera, respectively. Each landmark is manually annotated with a binary mask, highlighted in red color. The binary masks of two or more landmarks together with the corresponding uncalibrated disparity images are used to calibrate the target reference depth image, as described in section \ref{sec:calibration}.}
    \label{fig:reference_examples}
\end{figure}

\revb{
The data for this study was collected in the conservation area `Hintenteiche bei Biesenbrow' located in the Biosphere Reserve Schorfheide-Chorin. The data material is comprised of videos from 29 transects, captured using Bushnell Trophy CAM HD Agressor 119876 camera traps. The videos contain either greyscale infrared frames (captured at nighttime) or RGB (red, green, blue) color frames at 30 frames per second and a resolution of $1920 \times 1080 \text{px}$. We refer to greyscale and RGB images as intensity images. For each transect, a sequence of $N$ reference intensity images $\mathbf{I}^{\mathrm{ref}}_{i}$, with $i \in \{ 1, 2, ..., N\}$ were sampled manually from designated reference videos. Every such image shows a landmark with a known distance to the camera, in distances of 1 meter, 2 meters, ..., $N$ meters. These landmarks are established by a person showing a paper sheet depicting the distance to the camera by the number of meters. Figure \ref{fig:reference_examples} depicts two reference images with the researcher and the paper sheet positioned at a distance of 3 meters and 15 meters with respect to the camera, respectively. From the videos depicting observed animals, we automatically sample a single image every two seconds, from which both the manual as well as the automated distance measurements are derived. We refer to those images as observation images. To ensure no negative impact on further processing, we further remove metadata embedded visually inside each reference and observation image by cropping the bottommost $80 \text{px}$, reducing the effective resolution to $1920 \times 1000 \text{px}$. We exclude five out of the 29 total transects (T03, T04, T07, T11, T12) which were set up in a suboptimal way (c.f. section \ref{sec:guidelines}), which led to poor results. Hence, we do not include these transects in our evaluation.
}

Table \ref{tab:data_distribution} shows the distribution of reference and observation images with respect to the transects.

\begin{table}[H]
\resizebox{\textwidth}{!}{
\begin{tabular}{@{}lrrrrrrrrrrrr@{}}
\toprule
Transect       & \multicolumn{1}{c}{T01} & \multicolumn{1}{c}{T02} & \multicolumn{1}{c}{T05} & \multicolumn{1}{c}{T06} & \multicolumn{1}{c}{T08} & \multicolumn{1}{c}{T09} & \multicolumn{1}{c}{T10} & \multicolumn{1}{c}{T13} & \multicolumn{1}{c}{T14} & \multicolumn{1}{c}{T15} & \multicolumn{1}{c}{T16} & \multicolumn{1}{c}{T17} \\ \midrule
\# Ref. Images & 7                       & 7                       & 11                      & 14                      & 12                      & 13                      & 4                       & 9                       & 10                      & 10                      & 5                       & 5                       \\
\# Obs. Images & 4589                    & 5753                    & 920                     & 925                     & 942                     & 1220                    & 3246                    & 1949                    & 769                     & 886                     & 140                     & 59                      \\ \toprule
Transect       & \multicolumn{1}{c}{T18} & \multicolumn{1}{c}{T19} & \multicolumn{1}{c}{T20} & \multicolumn{1}{c}{T21} & \multicolumn{1}{c}{T22} & \multicolumn{1}{c}{T23} & \multicolumn{1}{c}{T24} & \multicolumn{1}{c}{T25} & \multicolumn{1}{c}{T26} & \multicolumn{1}{c}{T27} & \multicolumn{1}{c}{T28} & \multicolumn{1}{c}{T30} \\ \midrule
\# Ref. Images & 12                      & 15                      & 6                       & 6                       & 7                       & 10                      & 10                      & 15                      & 10                      & 7                       & 13                      & 13                      \\
\# Obs. Images & 160                     & 1111                    & 549                     & 5135                    & 425                     & 1210                    & 299                     & 422                     & 332                     & 125                     & 8356                    & 279                     \\ \bottomrule
\end{tabular}
}
\caption{Distribution of reference and observation images over \revi{the 24 considered transects}. The total number of observation images is \revi{$\SI{39801}{}$}. Reference images are used to determine the scale of the transect and observation images depict the animals to which the distance should be estimated.}
\label{tab:data_distribution}
\end{table}

\section{Methods}\label{sec:methods}

The challenge to overcome the distance estimation bottleneck \revi{in abundance estimation of unmarked animal populations} with simple monocular cameras, is the derivation of precise distance estimations to objects in the observed scene from just one single image.

Recent developments have shown that detailed distance estimations can be derived from a single image in an end-to-end manner based on deep learning approaches \citep{CAM-Convs}. Meanwhile, various deep learning have shown their effectiveness to address the monocular depth estimation (MDE). In this study, we decide for the DPT (Dense Prediction Transformers) approach that has shown superior quantitative and qualitative results in MDE. This is achieved by training on millions of pairs of monocular camera images and the corresponding distance estimations for each pixel \citep{midas,dpt}. The strength of DPT stems from employing a wide variety of training data from multiple sources.

\subsection{Camera trap calibration}\label{sec:calibration}

\begin{figure}
    \centering
    \begin{tikzpicture}
        \node[anchor=south west,inner sep=0] at (0,0) {\includegraphics[width=\textwidth]{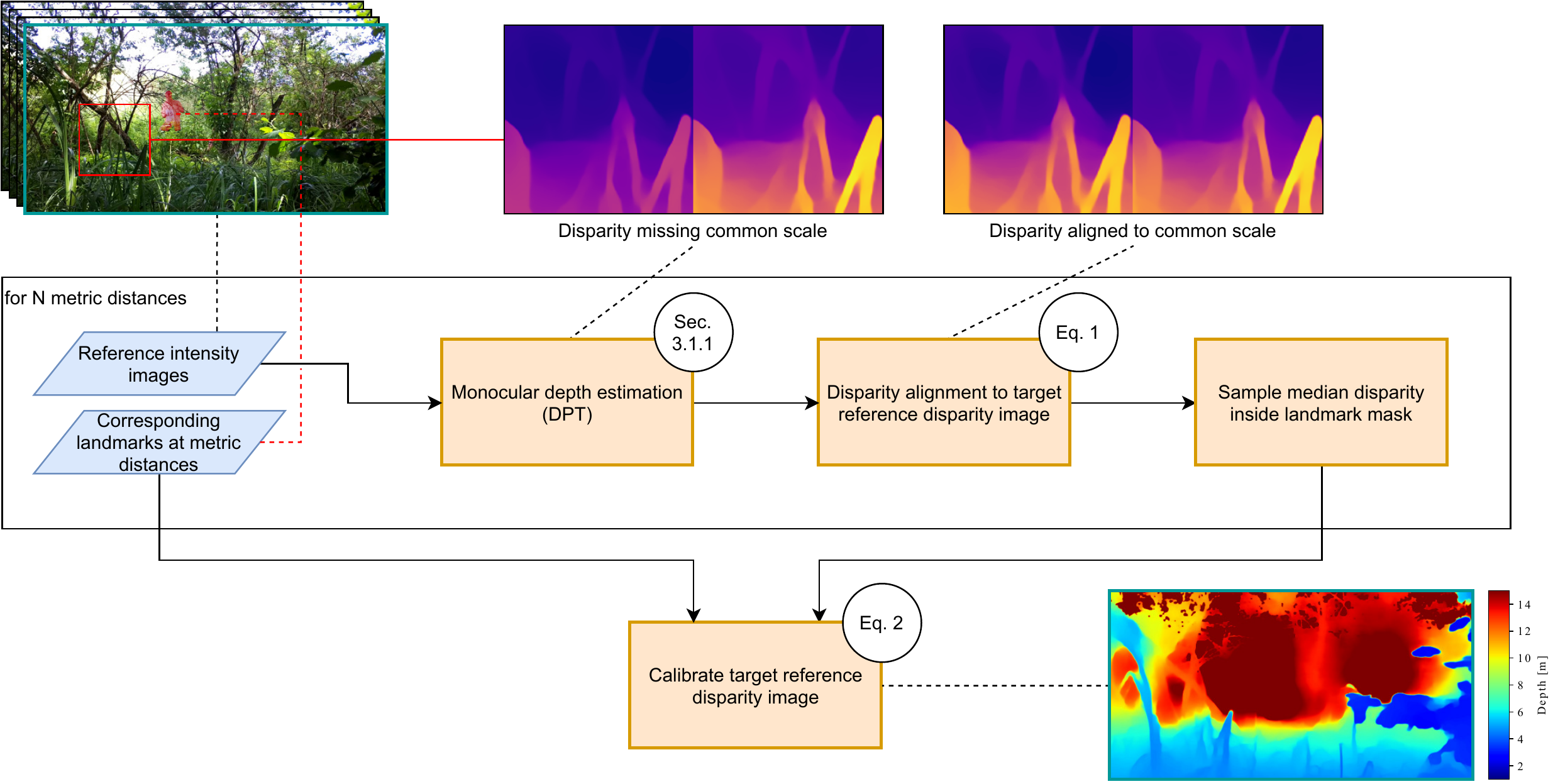}};
    \end{tikzpicture}
    \caption{Calibration workflow, which is processed once per transect. The $N$ reference intensity images are used to estimate $N$ corresponding uncalibrated disparity images, as described in section \ref{sec:mde)}. These uncalibrated disparity images (top-center image pair) are then aligned to a common scale (top-right image pair, c.f. equation \ref{eq:scale_alignment_ransac}). The median disparity value inside each landmark binary mask is used together with the known metric landmark distances to calibrate the single target reference disparity image (highlighted in blue, c.f. equation \ref{eq:calibration_ransac}). The target reference image is the reference image with the largest landmark distance.}
    \label{fig:workflow_calibration}
\end{figure}

Calibration of a camera trap employs reference images that depict landmarks of known distances to the camera. It is important to note that the reference images may be acquired in a multitude of ways since our calibration method is agnostic to the exact generation of the reference images. In this study, the landmarks are established by a person showing a paper sheet depicting the distance to the camera by the number of meters (cf. fig. \ref{fig:reference_examples}).

\subsubsection{Uncalibrated depth images via monocular depth estimation} 
\label{sec:mde)}
For each camera trap, there are $N$ reference images $\mathbf{I}^{\mathrm{ref}}_{i}$ with a corresponding binary mask $\mathbf{M}^{\mathrm{ref}}_{i}$ covering the landmark (depicted red in fig. \ref{fig:reference_examples}) and the corresponding true distance $z_i$ between camera and landmark for $i \in \{ 1, ..., N\}$. We refer to the $N$-th reference image as the \textit{target reference image}. The $N$ reference images $\mathbf{I}^{\mathrm{ref}}_{i}$ are first propagated through the DPT \citep{dpt} depth estimation model which results in $N$ uncalibrated disparity images $\mathbf{D}^{\mathrm{ref}}_{i}$ depicting the pixel-wise inverse distances to scene objects in a relative way, i.e., depth image pixels in blue are closer than those in green that in turn are closer than those in yellow which in turn are closer than those in red. More precisely: the uncalibrated disparity images show inverse distance estimations up to an unknown scale parameter $m$ and an unknown shift parameter $c$.

\subsubsection{Calibrated depth images via RANSAC}
Therefore, at least two landmarks with known distances to the camera must be used to determine both parameters. In this dataset, each of the $N$ reference images $\mathbf{I}^{\mathrm{ref}}_{i}$ depicts exactly one landmark, i.e, the researcher with a paper sheet.
Since the landmarks are distributed over all $N$ reference images $\mathbf{I}^{\mathrm{ref}}_{i}$, prior to the metric calibration, we align all uncalibrated disparity images to one common, yet not calibrated, scale. To be precise, for each uncalibrated disparity image $\mathbf{D}^{\mathrm{ref}}_{i}$ with $i \in \{ 1, ..., N - 1\}$, we estimate two parameters $m_i^*, c_i^*$, using the RANSAC approach \citep{ransac}, such that
\begin{equation}\label{eq:scale_alignment_ransac}
(m_i^*, c_i^*) \approx \argmin_{m_i, c_i} \sum_{i=1}^{N-1} |m_i \cdot \mathbf{D}^{\mathrm{ref}}_{i}(\mathbf{M}^{\mathrm{ref}}_{i}=0) + c_i - \mathbf{D}^{\mathrm{ref}}_{N}(\mathbf{M}^{\mathrm{ref}}_{N}=0)|,
\end{equation}
where $\mathbf{D}^{\mathrm{ref}}_{i}(\mathbf{M}^{\mathrm{ref}}_{i}=0)$ depicts all pixels in the disparity image $\mathbf{D}^{\mathrm{ref}}_{i}$ outside the binary mask $\mathbf{M}^{\mathrm{ref}}_{i}$ covering the landmark, i. e., all pixels depicting the visible stationary components of the observed scene. This alignment ensures the optimal alignment of all landmarks used in the next calibration step. Given the $N$ landmarks in the aligned uncalibrated disparity images $\mathbf{D}^{\mathrm{ref}}_{i}$ with $i \in \{ 1, ..., N - 1\}$, the RANSAC approach \citep{ransac} is then used to estimate the unknown scale parameter $m$ and the unknown shift parameter $c$ with the objective of minimizing the absolute disparity error:
\begin{equation}\label{eq:calibration_ransac}
(m^*, c^*) \approx \argmin_{m, c} \sum_{i=1}^{N} |m \cdot \mathrm{median}\left(m_i^* \cdot \mathbf{D}^{\mathrm{ref}}_{i}(\mathbf{M}^{\mathrm{ref}}_{i}=1) + c_i^*\right) + c - \frac{1}{z_i}|,
\end{equation}
where $\mathbf{D}^{\mathrm{ref}}_{i}(\mathbf{M}^{\mathrm{ref}}_{i}=1)$ depicts all pixels in the disparity image $\mathbf{D}^{\mathrm{ref}}_{i}$ within the binary mask $\mathbf{M}^{\mathrm{ref}}_{i}$ covering the landmark. From these disparity values the median value is chosen for minimization due to the improved robustness when facing imperfect landmark masks compared to the mean.
The real metric distance to the respective landmark (i.e., the ground truth) is depicted with $z_i$, the shift and scale parameters of disparity image $\mathbf{D}^{\mathrm{ref}}_{N}$ are given as $m_N^* = 1$ and $c_N^* = 0$. 
The resulting calibrated disparity images $\mathbf{C}^{\mathrm{ref}}_{i}$ and metric depth images $\mathbf{Z}^{\mathrm{ref}}_{i}$ are then given by equations \ref{eq:Cref} and \ref{eq:Zref}, respectively:
\begin{equation}
\label{eq:Cref}
\mathbf{C}^{\mathrm{ref}}_{i}=m^* \cdot \mathbf{D}^{\mathrm{ref}}_{i} + c^*,
\end{equation}
\begin{equation}
\label{eq:Zref}
\mathbf{Z}^{\mathrm{ref}}_{N}=(\mathbf{C}^{\mathrm{ref}}_{N})^{-1}.
\end{equation}

Instead of metric distances with image values in $\left[0, \infty\right]$ we can deal with disparity values in $\left[0, w\right]$ where $w$ is the image width. This results in improved numerical stability and induces lower weighting of more distant landmarks, reflecting the lower accuracy of the depth estimation at large distances. We refer to $\mathbf{C}^{\mathrm{ref}}_{N}$ and $\mathbf{Z}^{\mathrm{ref}}_{N}$ as the \textit{target} reference disparity and depth images, respectively. This target reference depth image is highlighted in blue in figure \ref{fig:workflow_calibration}.

\subsection{Animal distance estimation}\label{sec:animal_distance_estimation}

\begin{figure}
    \centering
    \begin{tikzpicture}
        \node[anchor=south west,inner sep=0] at (0,0) {\includegraphics[width=\textwidth]{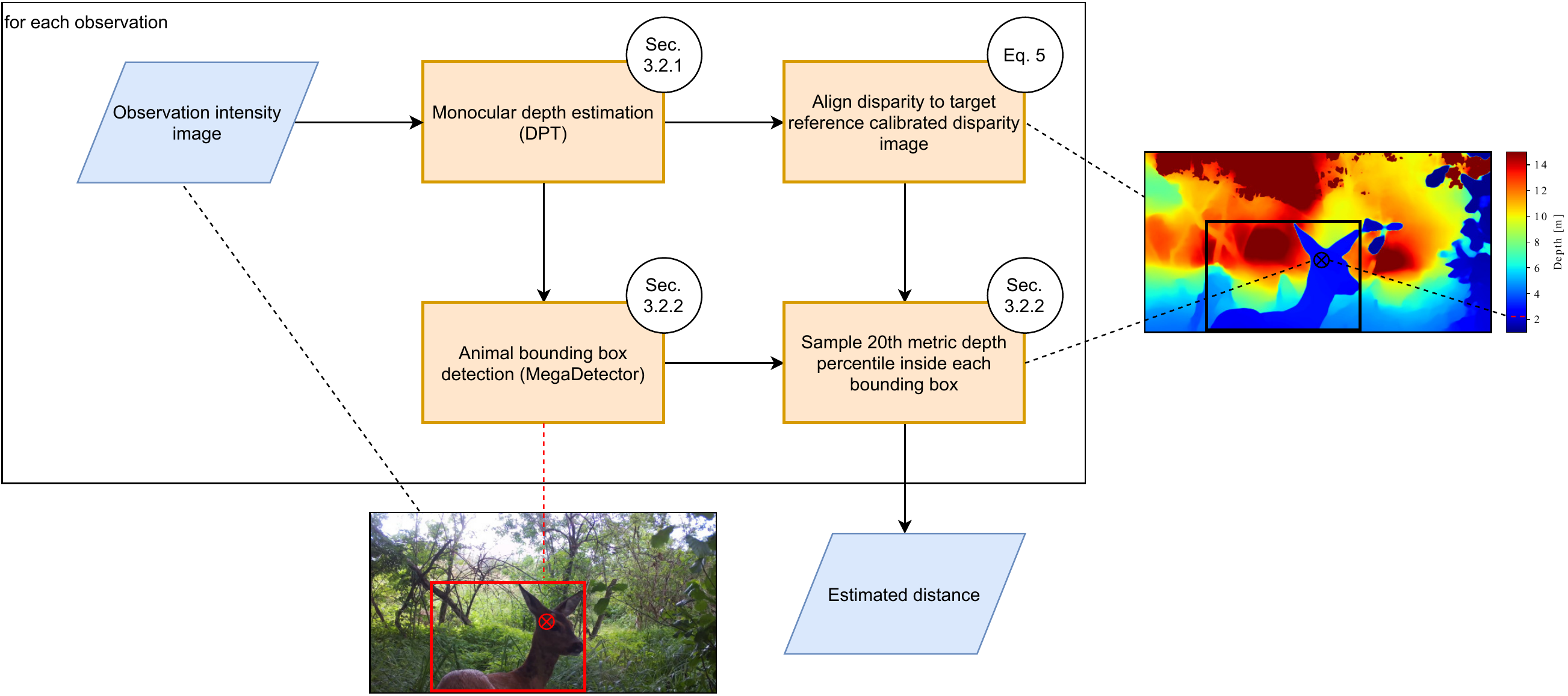}};
    \end{tikzpicture}
    \caption{Workflow which is applied on each animal observation image. From the intensity image, we estimate an uncalibrated disparity image, as described in section \ref{sec:depth_alignment}. We subsequently calibrate the observation depth image by aligning the depth to the target reference depth image (c.f. section \ref{eq:alignment_ransac}). We then sample the 20th percentile of the calibrated depth inside each detected animal bounding box to produce a single depth estimation for each animal (c.f. section \ref{sec:localization}).}
    \label{fig:workflow_observation}
\end{figure}

For each detected animal observation, we have to estimate a single metric distance to the animal. This objective demands to solve two requirements: (1) deriving a calibrated depth image $\mathbf{Z}^\mathrm{obs}$ of the camera trap image $\mathbf{I}^{\mathrm{obs}}$ depicting the observed animal, (2) localization of the observed animal in this calibrated depth image $\mathbf{Z}^\mathrm{obs}$.

\subsubsection{Deriving a calibrated depth image for each animal observation}\label{sec:depth_alignment}

Sampling accurate distance information for each observation image $\mathbf{I}^{\mathrm{obs}}$ employs the scale information of the calibration step described in section \ref{sec:calibration}. We achieve this by transferring the scale of the calibrated reference disparity images $\mathbf{C}^{\mathrm{ref}}_{i}$ to the estimated disparity images $\mathbf{D}^{\mathrm{obs}}$ of each animal observation. One might think that a simpler approach would be to just sample the depth of the calibrated reference images. However, the scenes observed by the camera traps are highly dynamic (due to trees falling over, plants gaining or loosing leaves, etc.), leading to higher estimation errors when employing this strategy. Therefore, we employ again the monocular depth estimation by DPT \cite{dpt} to estimate first an uncalibrated disparity image $\mathbf{D}^{\mathrm{obs}}$ of each observation image $\mathbf{I}^{\mathrm{obs}}$. We then transfer the metric scale acquired during calibration onto the uncalibrated disparity of each observation $\mathbf{D}^{\mathrm{obs}}$. From all possible $N$ calibrated reference disparity images $\mathbf{C}^{\mathrm{ref}}_{i},  i \in \{ 1, ..., N\}$ to inform this metric scale we use the calibrated target reference disparity $\mathbf{C}^{\mathrm{ref}}_{N}$, i.e., the one representing the calibration landmark with the largest distance. This choice shows the minimum number of pixels depicting the calibration landmark and therefore the maximum number of image pixels with an associated depth value that depict the scene where the animal is observed. We transfer the scale of the target depth image to the uncalibrated observation disparity image by again estimating the scale and shift parameters $m$ and $c$ using RANSAC \citep{ransac} while minimizing the absolute disparity error over the entire images, while excluding the calibration landmark and bounding boxes of detected animals (c.f. section \ref{sec:localization}):

\begin{equation}\label{eq:alignment_ransac}
(m^*, c^*) \approx \argmin_{m, c} |m \cdot \mathbf{C}^{\mathrm{ref}}_{N}(\mathbf{M}^{\mathrm{ref}}_{N}=0) + c - \mathbf{D}^{\mathrm{obs}}_{N}(\mathbf{M}^{\mathrm{obs}}_{N}=0)|
\end{equation}

Analogous to equation \ref{eq:Zref}, the result is the calibrated depth observation image $\mathbf{Z}^{\mathrm{obs}}$ of the observation image $\mathbf{I}^{\mathrm{obs}}$. \revi{This workflow is visualized by figure} \ref{fig:workflow_observation}.

\subsubsection{Localization of the observed animal in this calibrated depth image}\label{sec:localization}
\begin{figure}[H]
  \begin{center}
    \includegraphics[width=0.5\textwidth]{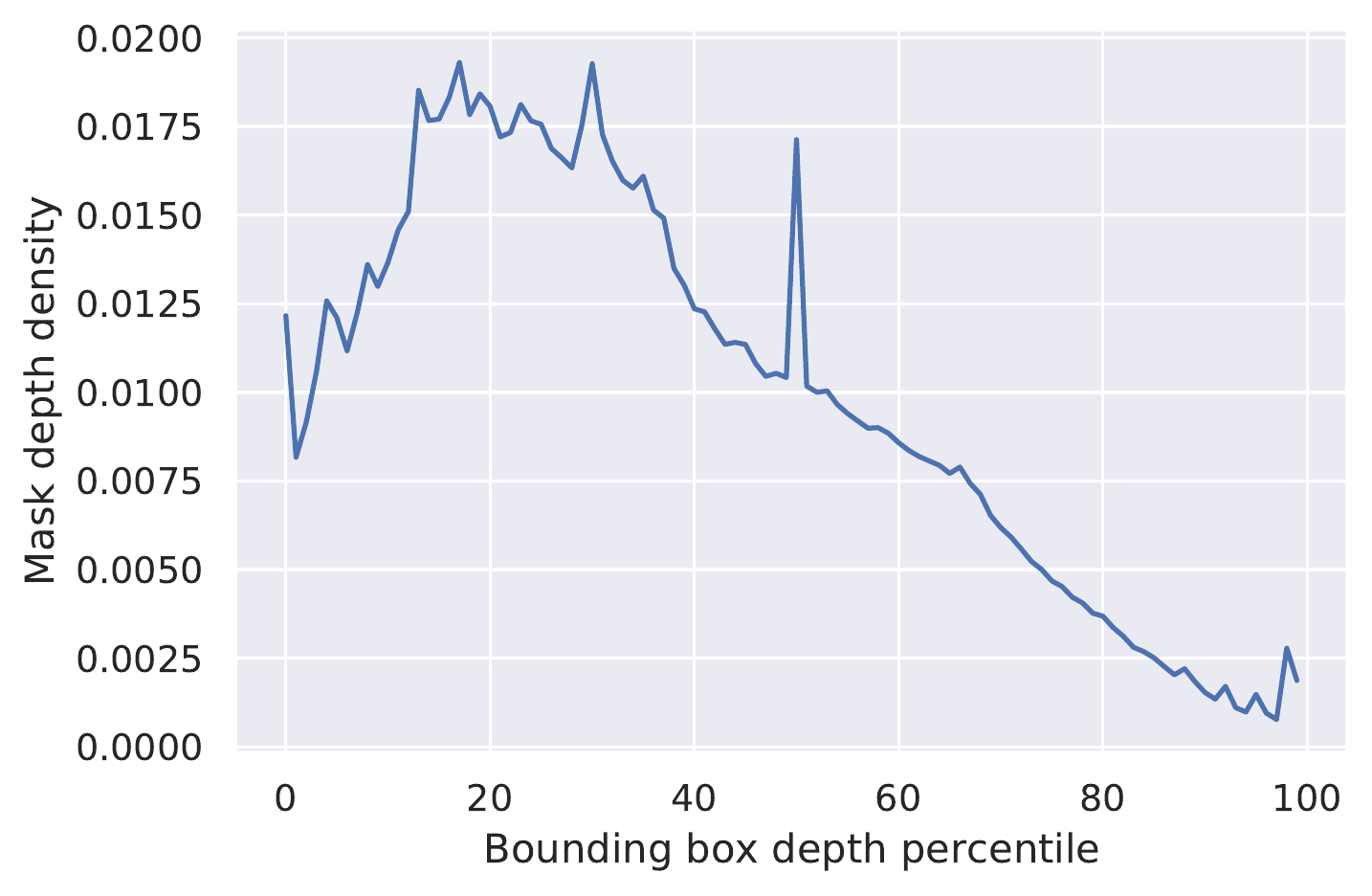}
  \end{center}
  \caption{Average density of depth inside manually annotated masks of 100 randomly sampled observations over the percentiles of depth values inside the enclosing bounding boxes. When sampling roughly at the 20th percentile of the depth contained inside the detected bounding boxes, the probability is maximal that the sampled depth is inside the manually annotated mask and therefore lies directly on the detected animal.}
  \label{fig:percentile_sampling}
\end{figure}

\begin{figure}[H]
     \centering
     \begin{subfigure}[b]{\textwidth}
         \centering
         \includegraphics[width=\textwidth]{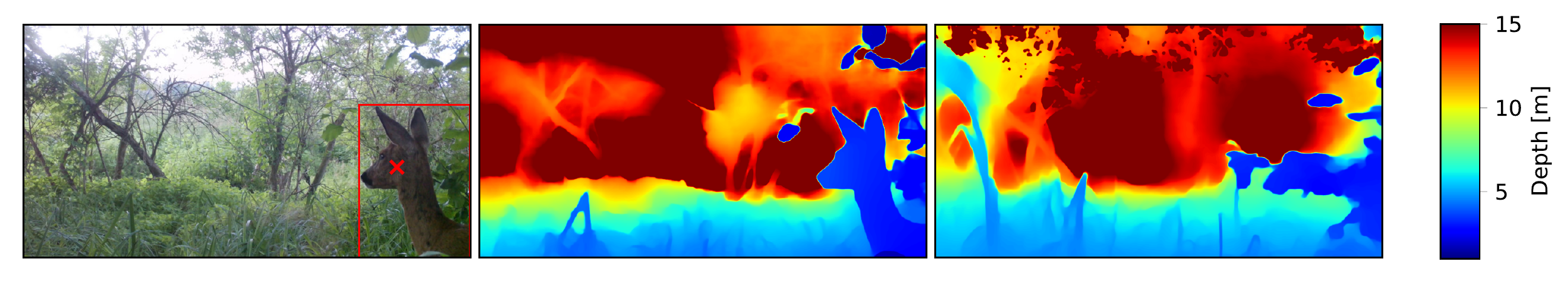}
         % T06___HT_cam10_433011_5887172_20190621___06210053___2019-06-21-21-22-45
         \label{fig:detection_example_1}
     \end{subfigure}
     \hfill
     \begin{subfigure}[b]{\textwidth}
         \centering
         \includegraphics[width=\textwidth]{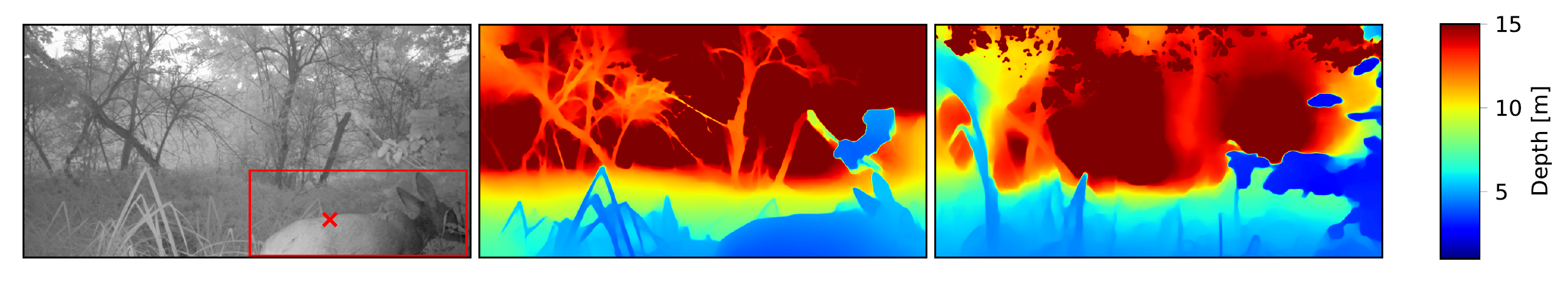}
         % T06___HT_cam10_433011_5887172_20190804___08040016___2019-08-04-21-09-20
         \label{fig:detection_example_2}
     \end{subfigure}
     \hfill
    \caption{Examples of a calibrated animal observation image. Left: Color (daytime) or infrared (nighttime) observation images with a bounding box and the \revi{resulting sampled distance} via the 20th percentile. Center: The corresponding estimated and calibrated depth image. Right: The target reference depth image corresponding to the calibration landmark with the largest distance. As can be observed, the background changes slightly between both images. This is due to the fact that both images were captured with a difference in time of two months.}
    \label{fig:detection_examples}
\end{figure}

\revb{
For animal detection we employ MegaDetector \citep{megadetector}, a deep-learning animal detection model based on the Faster R-CNN \citep{fasterrcnn} and Inception Resnet \citep{inceptionresnet} architecture. It is trained using large amounts of images annotated by humans with bounding boxes for the object classes \textit{animal}, \textit{human}, and \textit{vehicle}. We use this trained MegaDetector model and apply it to the observation image $\mathbf{I}^{\mathrm{obs}}$, resulting in a bounding box for each animal observed in $\mathbf{I}^{\mathrm{obs}}$. From all detected bounding boxes corresponding to a single observation, we infer a binary mask $\mathbf{M}^{\mathrm{obs}}$ which is set to one at each pixel inside any detected bounding box and to zero everywhere else. This binary mask is used in equation \ref{eq:alignment_ransac}. Then, we sample for each bounding the 20th percentile of the corresponding calibrated depth observation image $\mathbf{Z}^{\mathrm{obs}}$. Figure \ref{fig:detection_examples} shows two exemplary observation images with corresponding detected bounding boxes, depth images and the locations of the sampled depth.
}
This procedure is simple but effective. It is also intuitive, as the animals are mostly positioned on a much more distant background and slightly occluded by plants or trees. The 20th percentile of the depth then presents an accurate estimate of the true distance, as illustrated by figure \ref{fig:percentile_sampling}.
We also evaluated more sophisticated methods for precise localization such as class attention maps (CAMs, \cite{cam}) of species classification models \citep{ms-species-classificator} but found these models to fail in many instances when the animals are strongly occluded. The classification of animals is therefore performed by a human observer and not automated.

\subsubsection{\revi{Metrics for evaluating measurement error}}\label{sec:metrics}
For evaluation, we employ the mean absolute distance estimation error over all observations $m \in \{ 1, ..., M\}$, defined as:
\begin{equation}
    \frac{1}{M}\sum_{m=1}^{M}\left|z_{m}^{\mathrm{est}}-z_{m}^\mathrm{gt}\right|
\end{equation}
and the mean distance estimation error in our evaluation, defined as:
\revb{%
\begin{equation}
    \frac{1}{M}\sum_{m=1}^{M}\left(z_{m}^{\mathrm{est}}-z_{m}^\mathrm{gt}\right)
\end{equation}
}%
where $z_{m}^{\mathrm{est}}$ and $z_{m}^\mathrm{gt}$ represent the estimated and ground-truth distance of each observation, respectively.

\subsection{\revi{Distance Estimation Workbench}}
\revb{%
We implement the above methodology using the Python programming language. The execution of the MegaDetector and DPT models is handled by the TensorFlow \citep{tensorflow} and PyTorch \citep{pytorch} libraries, respectively. The RANSAC \citep{ransac} implementation is provided by Scikit-learn \citep{sklearn}. To make our methodology available to other researchers, we provide it in the form of a simple graphical user interface, which we call \textit{Distance Estimation Workbench}. The Distance Estimation Workbench allows starting and stopping individual parts of the calibration and distance estimation workflows. Input and output is follows a standardized directory structure for image data and CSV spreadsheet files are used for metadata and ground truth measurements. This allows efficient processing of large datasets without manual interaction. The executable Distance Estimation Workbench is available, together with accompanying documentation and a minimal example dataset, at: \href{https://timm.haucke.xyz/publications/distance-estimation-animal-abundance}{https://timm.haucke.xyz/publications/distance-estimation-animal-abundance}
}%

\section{Evaluation and discussion}\label{sec:eval}
\begin{figure}
    \centering
    \includegraphics[width=\textwidth]{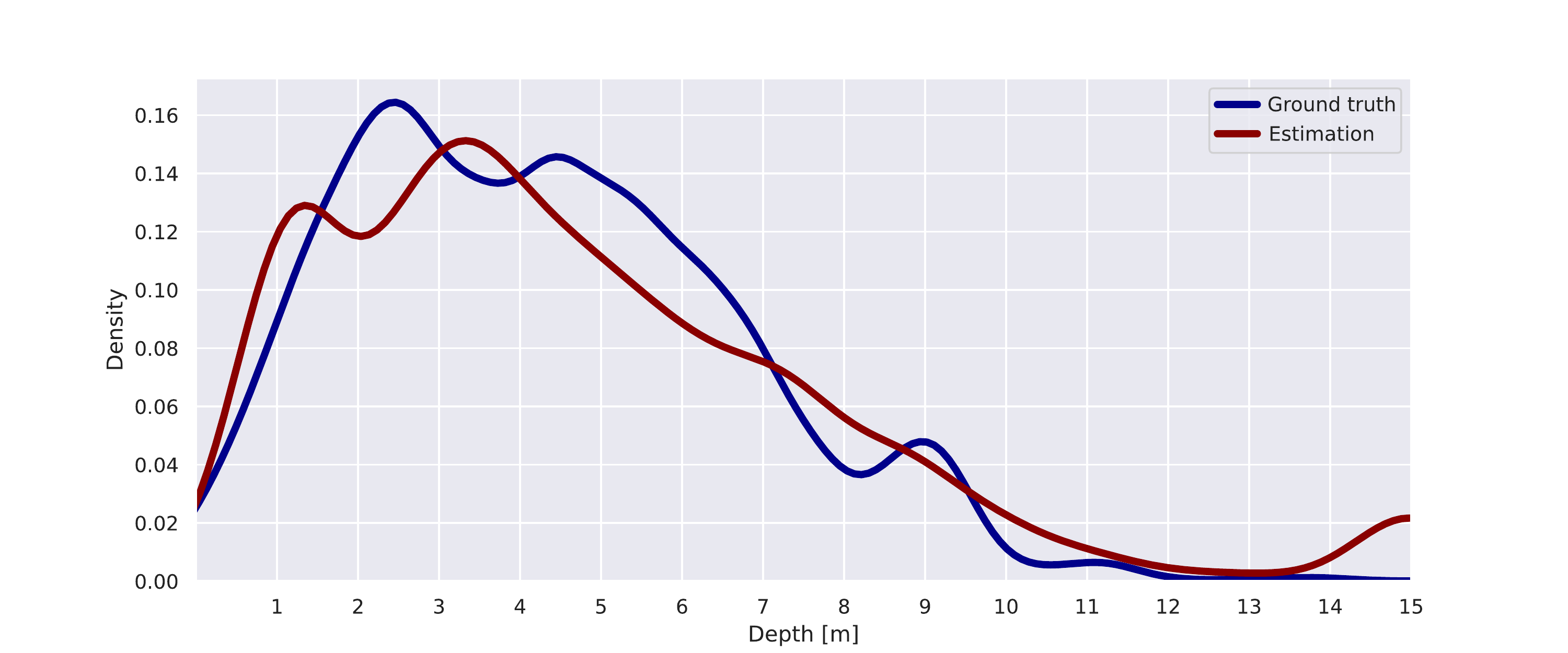}
    \caption{Probability density of the ground truth and estimated distances obtained using kernel density estimation. As can be seen, the distribution of estimated distances closely matches the ground truth distance distribution.}
    \label{fig:density}
\end{figure}

\begin{figure}[H]
    \centering
    \includegraphics[width=1\textwidth]{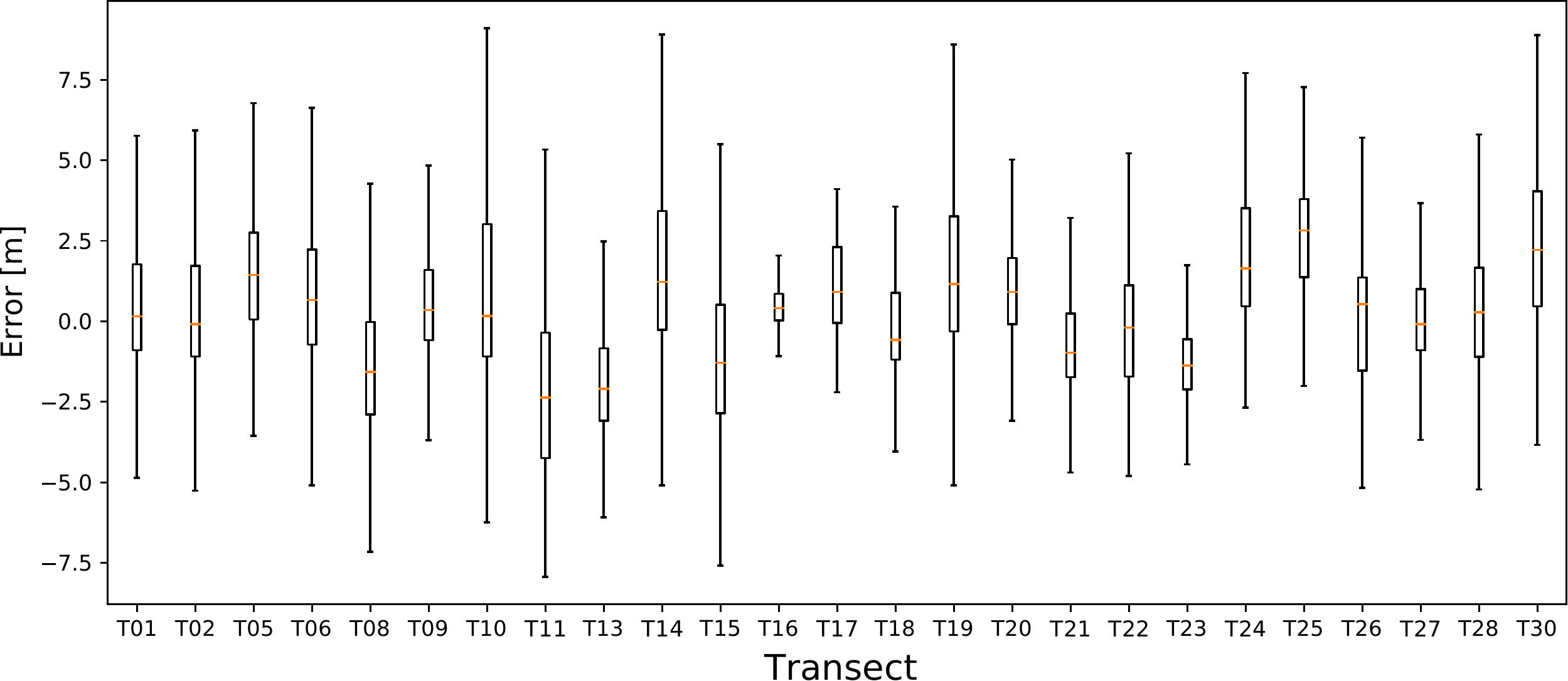}
    \caption{Box plot of the distance estimation error per transect.}
    \label{fig:transect_error_boxplot}
\end{figure}

For the resulting distance estimations to be usable for \revi{the various methods available for the estimation of abundance of unmarked animal populations}, it is important that our estimation method produces a distance distribution as close to the ground truth and as unbiased as possible. As can be seen in figure \ref{fig:density}, the distribution of estimated distances indeed reflects the ground truth distribution. At $\SI{2}{\metre}$ both distributions differ by about 4 percentage points while the difference at $\SI{9}{\metre}$ is about 1 percentage point.
We achieve a mean distance error of $\SI{0.10}{\metre}$ and a mean absolute distance error of $\SI{1.85}{\metre}$. The small positive bias of our method can be explained by the distribution of distance values in the calibrated depth images. Large parts of the depth images show background areas with arbitrarily large distances. If an animal is falsely detected in such an area, a very large distance is falsely estimated. Both the mean and the mean absolute distance error measures depend strongly on the transect, as can be seen in figure \ref{fig:transect_error_boxplot}. High estimation errors can be observed with dense vegetation directly in front of the camera (e.g., T24), as the employed monocular depth estimation tends to smooth out the estimated disparity images, which is especially damaging for small cavities in the vegetation, in which the background then appears closer than it truly is. In this case, the initial calibration (c.f. section \ref{sec:calibration}) fails because the known landmarks appear to be in a single plane. In other transects (e.g., T02), the forest ground is only visible to a small degree. This apparently also reduces monocular depth estimation accuracy because important context information about the relative location of objects in the scene is lost.

\subsection{Camera trap setup guidelines}\label{sec:guidelines}
\begin{figure}[H]
     \centering
     \begin{subfigure}[b]{\textwidth}
         \centering
         \includegraphics[width=0.98\textwidth]{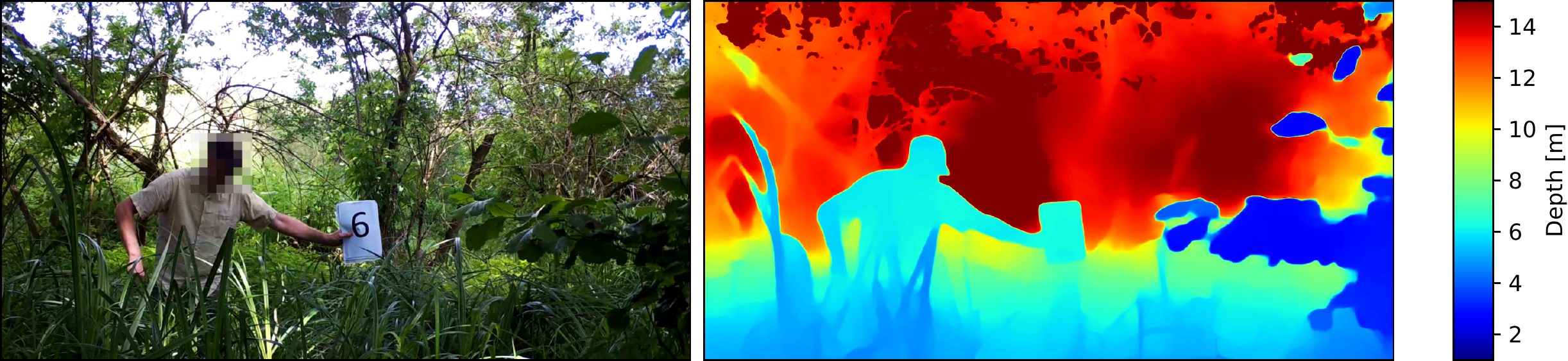}
     \end{subfigure}
     \hfill
    \caption{Exemplary reference image of transect T06. Left: Color (daytime) reference image. Right: The corresponding calibrated reference depth image obtained as described in section \ref{sec:calibration}.}
    \label{fig:calibration_examples}
\end{figure}

The choice of scene and the camera setup is therefore an important factor for the success of our method. A calibration result of a well-conditioned setup can be seen in figure \ref{fig:calibration_examples}. We want to provide researchers with guidelines on where and how to best place camera traps in the future to make the best use of our method and therefore make the following recommendations:
\begin{itemize}
    \item Camera traps should be tightly secured to stationary objects, i.e. trees. This reduces camera motion and hence ensures a strong overlap of observation images with reference images
    \item generally, camera trapping benefits from a free field of view, therefore it should be free of vegetation inside a radius of three meters
    \item at least the bottom third of the image produced by the camera trap should be covered by the ground to ensure enough context information for the monocular depth estimation
    % TODO
    % \item if possible, in the future, standardized reference objects \revi{(e.g. Fluchtstäbe)} should be placed and stay in the scene so that accurate calibration can be performed separately for each observation, without the now required reference images. \revb{Using standardized reference objects has the benefit of enabling the automated detection thereof via conventional computer vision techniques such as template matching. If this is not desirable, e.g. due to conservation interests, fixed objects such as trees, big rocks or logs could be used instead}
    \item \revb{if the situation allows, artificial (e.g. ranging rods) or natural (e.g. trees, rocks, logs) \citep{palencia2021assessing} reference objects could be permanently placed in the scene and incorporated in our automated method. A minimum of two reference objects are required for our calibration workflow, however, the more reference objects are captured, the more robust the calibration becomes}
\end{itemize}

\subsection{\revi{Evaluating distance estimation effort}}\label{sec:userstudy}
To quantify the reduction of the manual distance estimation workload facilitated by our method, we conducted a user study with five users experienced with wildlife monitoring using camera trap imagery. Out of the data described in table \ref{tab:data_distribution}, we randomly chose five transects, out of which we randomly sampled five detection videos with no more than one single animal present at a time. The participants of the study are then asked to apply the manual distance estimation process (cf. appendix \ref{sec:manual_distance_estimation_process}). We chose only observations with at most a single animal present at a time to prevent ambiguous assignments between multiple individuals over the participants and to therefore be able to quantify the deviation of distance estimations between participants.
The time needed by a participant to compare the position of an observed animal in a video frame to the different distances in the reference video clips and estimate the distance has been measured to lie between $\SI{8.6}{\second}$ and $\SI{17.9}{\second}$. The mean time needed per observation is $\SI{12}{\second}$.
%
% \revi{We extrapolate to the complete dataset by multiplying the time needed per observation with the total number of observation images, resulting in a total effort of about 130 person hours. Some images in the complete dataset contain multiple animals and the effort to estimate distances for each need more time than for a single animal. A total effort of about 130 person hours is therefore a lower bound.}

\revi{We then estimate the workload of manual distance estimation of the complete dataset by assuming that every observation image shows only a single animal. Our comparison is therefore based on the processing time per observation image. This results in $130$ person hours for the complete dataset of $\SI{39801}{}$ observation images. However, about $4\%$ of the $\SI{39801}{}$ total observation images contain more than one animal. Therefore, the $130$ person hours slightly underestimate the manual distance estimation workload for the complete dataset by assuming that every observation image shows only a single animal.

Our automated distance estimation pipeline requires $6$ person hours for annotating $240$ reference images and 24 hours for automated distance estimation for all $\SI{39801}{}$ observation images. 

The ratio between the complete manual distance estimation effort (${\SI{130}{\hour}}$) and the complete automated distance estimation effort (${\SI{6}{\hour}}$ + ${\SI{24}{\hour}}$) is $\frac{\SI{130}{\hour}}{\SI{30}{\hour}} = 4.33$. Since the time required for the manual distance estimation is underestimated, this ratio of $4.33$ is a lower bound of the speedup factor. The same holds for the speedup factor of the purely manual workload, which is $\frac{\SI{130}{\hour}}{\SI{6}{\hour}} = 21.66$}.

We also compared the quality of the manual distance estimations produced by the participants. In 9\% of cases, the participants disagree on whether an animal is visible in the image. The mean standard deviation between the participants over the remaining 91\% of measurements is $\SI{62}{\centi\metre}$, suggesting a lower bound of the achievable accuracy.

\section{Conclusion}\label{sec:conclusion}
\revb{
Methods for abundance estimation of unmarked animal populations from camera traps all require an estimate of the effective area surveyed, which is usually done by deriving camera-to-animal observation distances. This is time-consuming, error-prone and subjective, which motivates our automated distance estimation method based on monocular depth estimation and a robust calibration workflow.
Our method imposes no constraints on specific camera hardware and is therefore applicable to a wide variety of datasets. In our experiments, we succeed in closely matching the true distance distribution.
Thereby we successfully overcome the distance estimation bottleneck in abundance estimation of unmarked animal populations. Our automated method achieves a mean distance error of only 0.14 m, it reduces the manual effort \revi{by a factor of $21.66$ and the total processing time by a factor of $4.33$}. This facilitates large-scale, automated abundance estimation of unmarked animal populations.}
\\[1ex]
Future work could improve the temporal stability of monocular depth estimation and in turn further improve the distance estimation accuracy.
\revb{%
In cases where videos or image sequences are available for each animal observation, multi-object tracking approaches would likely reduce false positive and false negative observations by combining information from multiple frames.
}

\newpage

\section*{Acknowledgments}
This research has been funded in part by the Federal Ministry of Education and Research (www.bmbf.de) of the Federal Republic of Germany under grant number 01DK17048. We would like to thank the Helversen'sche Stiftung for providing permission and access to the FFH conservation area `Hintenteiche bei Biesenbrow' and are grateful for support of our work by Dorothea Dietrich, Dietmar Nill, Ulrich Stöcker and Thomas Volpers. We thank Dr. Martin Flade and Rüdiger Michels from the Biosphere Reserve Schorfheide-Chorin. \revi{We thank the participants of the user study. We thank Frank Schindler for proofreading the manuscript.}

\clearpage

\begin{appendices}
\section{Manual distance estimation process}\label{sec:manual_distance_estimation_process}
\begin{algorithm}[H]
    \SetAlgoLined
    \DontPrintSemicolon
    \KwResult{Manual distance estimations}
    open a spreadsheet\;
    \ForEach{transect}{
        note the transect and starting time in the spreadsheet\;
        \ForEach{observation video in transect}{
            open the respective video file\;
            \ForEach{video runtime from $0$ to $\SI{58}{\second}$ in $\SI{2}{\second}$ steps}{
                pause the video \;
                locate the animal \;
                \eIf{an animal is present}{
                    compare the position of the animal to the different distances in the reference images\;
                    estimate the most accurate lower integer distance bound and note it in the spreadsheet\;
                }{
                    note that no animal is present\;
                }
            }
        }
        note the elapsed time in the spreadsheet\;
    }
    \caption{Description of the manual distance estimation process which we employed in the user study}
\end{algorithm}

\end{appendices}

\pagebreak

\Urlmuskip=0mu plus 1mu\relax
\bibliographystyle{elsarticle-harv}
\bibliography{references}

\end{document}